\pdfoutput=1

\documentclass[11pt]{article}

\usepackage{acl}

\usepackage{soul}
\usepackage{color}
\usepackage{mdframed}
\usepackage{times}
\usepackage{latexsym}
\usepackage{graphicx}
\usepackage[T1]{fontenc}
\usepackage[utf8]{inputenc}
\usepackage{booktabs}
\usepackage{microtype}
\usepackage{inconsolata}
\usepackage{multicol, multirow}
\usepackage{listings}
\usepackage{amsmath}
\usepackage{amssymb}
\usepackage{pifont}
\usepackage{adjustbox}
\newcommand{\xmark}{\ding{55}}
\newcommand{\cmark}{\ding{51}}
\newcommand{\paratopic}[1]{}

\newcommand*{\myfont}{\fontfamily{pcr}\selectfont}
\newcommand{\alert}[1]{#1} % <== turn off

\definecolor{lightred}{rgb}{0.929, 0.8, 0.8}
\definecolor{palegreen}{rgb}{0.812, 0.961, 0.843}
\definecolor{lightyellow}{rgb}{1, 1, 0.784}

\title{Claim Check-Worthiness Detection:\\How Well do LLMs Grasp Annotation Guidelines?}
\author{Laura Majer \and Jan Šnajder  \\ Text Analysis and Knowledge Engineering Lab \\ University of Zagreb, Faculty of Electrical Engineering and Computing \\ \texttt{\{laura.majer, jan.snajder\}@fer.hr}}

\begin{document}
\maketitle
\begin{abstract}

%While CD is commonly defined as a classification task, CW may also be framed as a ranking task, mirroring prioritization done by fact-checkers. 
%The starting point of the pipeline is claim detection (identifying text segments requirying fact-checking) and claim check-worthiness 
The rising threat of disinformation underscores the need to fully or partially automate the fact-checking process. Identifying text segments requiring fact-checking is known as \emph{claim detection} (CD) and \emph{claim check-worthiness detection} (CW), the latter incorporating complex domain-specific criteria of worthiness and often framed as a ranking task. Zero- and few-shot LLM prompting is an attractive option for both tasks, as it bypasses the need for labeled datasets and allows verbalized claim and worthiness criteria to be directly used for prompting. We evaluate the LLMs' predictive accuracy on five CD/CW datasets from diverse domains, using corresponding annotation guidelines in prompts. We examine two key aspects: (1) how to best distill factuality and worthiness criteria into a prompt, and (2) how much context to provide for each claim. To this end, we experiment with different levels of prompt verbosity and varying amounts of contextual information given to the model. We additionally evaluate the top-performing models with ranking metrics, resembling prioritization done by fact-checkers. Our results show that optimal prompt verbosity varies, meta-data alone adds more performance boost than co-text, and confidence scores can be directly used to produce reliable check-worthiness rankings.
\end{abstract}

\section {Introduction}

%\paratopic {Introduce automated FC, FC organisations, CD}

\begin{figure}[t]
\centering
% \hspace*{-0.8cm} 
\includegraphics[width=0.52\textwidth]{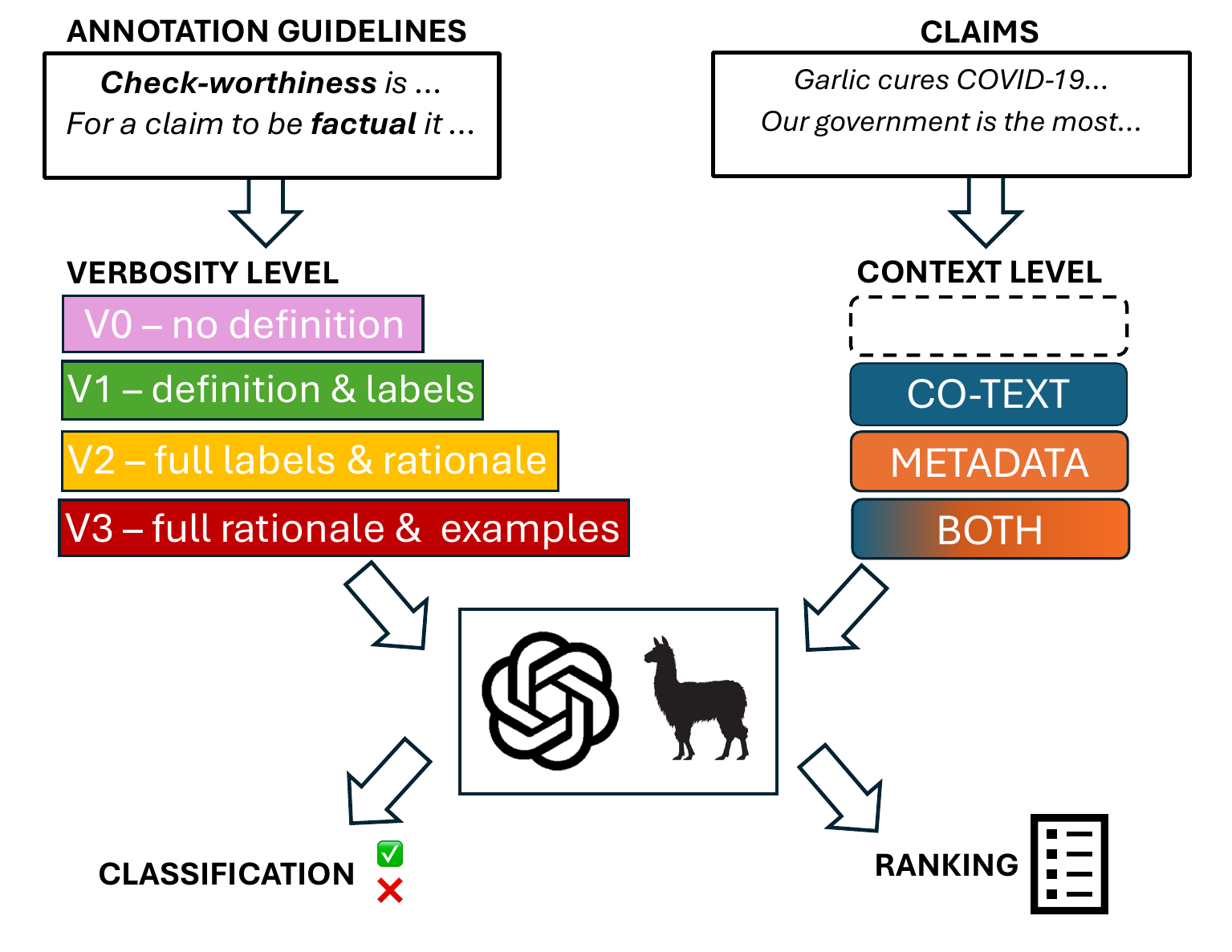}
\caption{Using annotation guidelines, we craft zero- and few-shot LLM prompts for claim and claim check-worthiness detection, varying the level of prompt verbosity and the amount of provided context. We evaluate the LLMs using classification and ranking metrics.}
\label{fig:info}
\end{figure}

The global spread of information, coupled with mis- and disinformation, is increasing the demand for fact-checking \citep{un2022disinfo, undp2023disinfo}, highlighting the need for automation. However, complete automation may not be ideal; for instance, PolitiFact, which used ChatGPT to verify previously fact-checked claims, faced issues like inconsistency, knowledge limitations, and misleading confidence \citep{poynter2023gpt}. Nevertheless, they see potential in language models for assisting fact-checkers, particularly in identifying claims worth verifying. Similarly, FullFact has highlighted the lack of effective claim selection tools as a major workflow challenge \citep{fullfact}.

%However, they recognize the potential of using language models to assist fact-checkers, especially in identifying claims worthy of verification. Similarly, FullFact, another fact-checking organization, identified the lack of effective tools for selecting claims to check as a primary workflow challenge \citep{fullfact}.

%\paratopic{Introduce CD and CW}

%Automating fact-checking is becoming crucial in response to rising amounts of data and disinformation. 
To warrant fact-checking, a claim must be both \emph{factual} (i.e., related to purported facts) and \emph{check-worthy} (i.e., of interest to society). The NLP tasks of identifying factual and check-worthy claims are known as \emph{claim detection} (CD) and \emph{claim check-worthiness detection} (CW), respectively. The tasks make up the first component of the automatic fact-checking pipeline. While both are typically defined as classification tasks, CW can also be framed as a ranking task, mimicking the prioritization process employed by fact-checking organizations. %\citep{fullfact}.

%\paratopic{The complexities of CD and CW}

Both CD and CW are challenging for several reasons. Firstly, the underlying concepts of factual claims and check-worthiness resist straightforward definitions. 
To grasp factuality, \citet{konstantinovskiy2021} presented a thorough categorization of factual claims, while \citet{ni2024afacta} provided a definition distinguishing opinions. Regardless of these variations, \emph{factual} could be deemed universal and self-explanatory, unlike \emph{check-worthiness}, a term frequently used in previous research.
Defining check-worthiness is made more challenging by its subjective, context-dependent nature and temporal variability.
Assessing it usually requires choosing more specific criteria, such as relevance to the general public \citep{hassan2017toward} or policymakers, potential harm \citep{CLEF2022}, or alignment with a particular topic \citep{stammbach-etal-2023-environmental, gangi-reddy-etal-2022-newsclaims}).
Another challenge is identifying the situational context (including previous discourse and speaker information) required to determine claim factuality and check-worthiness. For example, in CW annotation campaigns \citep{hassan2017toward, gangi-reddy-etal-2022-newsclaims}, annotators are typically presented with surrounding sentences to aid their assessment.

%In a definition given by \citet{ni2024afacta}, a factual claim is a statement that explicitly presents some verifiable facts. Additionally, statements with subjective components like opinions can also be factual claims if they explicitly present objectively verifiable facts.
%Defining check-worthiness is made more challenging by its subjective, context-dependent nature and temporal variability; 

%\maybe{\citet{gencheva-etal-2017-context} point to disagreement among fact-checking organizations regarding which claims to check even when provided the same input.}
%-- challenges that prompted \citep{konstantinovskiy2021} to conclude that it 'should be left to fact-checkers' . 

%\citet{gencheva-etal-2017-context} highlight the importance of contextual features based on surrounding sentences for claim detection, while \citet{gangi-reddy-etal-2022-newsclaims} argue for considering both the speaker and the claim object. 

%The boundaries between factual and check-worthy claims are blurry also during their detection. There are two options of differentiating the tasks. One option is treating them as sequential tasks -- after factual claims are selected, they are additionally scored on importance. Alternatively, importance can included in the task immediately, serving as a hard-pass filter.

%\paratopic{Solutions to this task}  

The CD and CW tasks have been approached using both traditional supervised machine learning and fine-tuning pre-trained language models, both of which depend on labeled data. However, obtaining such datasets can be challenging as they need to align with specific languages, domains, and genres and meet desired factuality and worthiness criteria. Moreover, dataset annotation is costly and requires redoing if criteria change.
%
%\paratopic{LLMs}
%
LLMs present a viable alternative to supervised methods owing to their strong zero- and few-shot performance \citep{NEURIPS2022_8bb0d291,NEURIPS2020_1457c0d6}.
%Due to their in-context learning abilities \citep{brown2020language}
%They are able to generalize and have strong zero-shot performance \citep{moskvoretskii2024taxollama}, as well as having broad world-knowledge. 
%Sophisticated LLM pipelines work well for CD \citep{ni2024afacta}, but CW is more %challenging due to its subjective nature. 
Over time, fact-checking organizations have refined principles for claim prioritization, and zero- and few-shot prompting offers a seamless way to transfer this knowledge to the model. Thus, an effective strategy might entail zero- and few-shot prompting with check-worthiness criteria from annotation guidelines.
The challenge, however, is that LLMs often exhibit sensitivity to variations in prompts \citep{mizrahi2024state} and unreliability \citep{si2023large}.

%Fact-checking organizations typically prioritize claims using structured systems, whether explicitly defined or evolving organically. Zero-shot prompting offers a direct way to transfer this knowledge to models, as worthiness can be expressed in natural language, making prompt creation similar to drafting guidelines but with a less labor intensive process that follows.

%Addressing CW involves refining 'worthiness criteria' to specific claim characteristics, which doesn't always lead to simplification, as it may encompass multiple requirements or unique categorization of claims. 

%we are investigating both the more complex task of CW and solving it in a more approachable way: easily-deployable zero-shot prompting. This method requires no prompt engineering expertise, making it approachable for fact-checking organisations.

%Since fact-checking organizations typically have a structured system for claim prioritization, whether explicitly established or organically evolving from their claim selection process, zero-shot prompting could be a straightforward method for transferring such knowledge to a model. Worthiness or priority can be articulated in natural language, therefore, constructing the prompt might be no different from drafting guidelines while being less time consuming and more economical.

\paratopic{In this paper}

In this paper, we study the predictive and calibration accuracy of zero- and few-shot LLM prompting for CD and CW. 
We experiment with five datasets, each with a different factuality or worthiness criterion outlined in the accompanying annotation guidelines. We investigate two key aspects: (1) how to best distill factuality and worthiness criteria from the annotation guidelines into the prompt and (2) what amount of context to provide for each claim. For (1), we experiment with varying the level of prompt verbosity, starting from brief zero-shot prompts to more detailed few-shot prompts that include examples. For (2), we expand the prompt with co-text and other components of the claim's situational context.
%
%To cover different domains, we experiment with five diverse datasets that provide annotation guidelines.
%We rely merely on annotation guidelines to build prompts. Since prompt engineering expertise is not a prerequisite for this method, it could potentially be useful in fact-checking organizations for automating CWD. Due to the complex nature of the tasks, a certain level of verbosity is needed to express all desired criteria. However, it is not clear to which extent this transfers to zero-shot inference. To investigate that, we vary the prompts in levels of verbosity, starting from brief zero-shot prompts to detailed ones including examples. 
%To recognise the significance of context for the task, we expand the prompt with co-text and other components of situational context. 
Furthermore, inspired by the fact-checker's prioritization process, we consider CW as a ranking task, using LLM confidence scores as a proxy for determining priority. Figure ~\ref{fig:info} depicts the workflow of our experiments. 
We show that prompting with worthiness criteria adopted from annotation guidelines can yield accuracy and ranking scores comparable to or surpassing existing CD/CW methods. Although optimal prompt verbosity varies across datasets, certain in-domain trends can be observed across models. We also find that the impact of adding context is greater for lower verbosity levels, while meta-data is more beneficial than co-text. Finally, we show that confidence scores can be directly used to produce reliable check-worthiness rankings.
%
%evaluate the ranking select the optimal prompt variant based on predictive and calibration accuracy and calculate ranking metrics using confidence as a proxy for priority.
%
%\paratopic{Contributions+impact}

Our contributions include analyzing LLM performance in terms of (1) prompt detail, (2) provided context, and (3) variations across domains and worthiness criteria.

\section{Related Work}

Developing a fully automated fact-checking system is appealing for both its applicability and the challenge it presents \citep{hassanCB, li2023selfchecker}. However, \citet{glockner-etal-2022-missing} question the purpose of such a system, pointing to its reliance on counter-evidence that may not be available for newly coined disinformation. This motivates a shift toward human-in-the-loop approaches and automating parts of the fact-checking pipeline.

The CD and CW tasks constitute the first part of the fact-checking pipeline and are meant to select parts of the input for which fact-checking is possible (CD) or deemed necessary (CW). Typically framed as classification tasks, the CD and CW tasks are handled using traditional supervised machine learning \citep{Hassan2017TowardAF, wright-augenstein-2020-claim, hassan2017toward, gencheva-etal-2017-context} or fine-tuning pre-trained language models \citep{stammbach-etal-2023-environmental, sheikhi-etal-2023-automated}. 
Methods of solving include rich sentence and context-level features \citep{gencheva-etal-2017-context}, speaker, object, and claim span identification \citep{gangi-reddy-etal-2022-newsclaims}, or incorporating domain-specific knowledge by combining ontology and sentence embeddings \citep{hüsünbeyi2024ontology}. CW can also be framed as a ranking task \citep{jaradat-etal-2018-claimrank, gencheva-etal-2017-context}, mimicking the prioritization of claims by fact-checking organizations. 

Recently, the use of LLMs for CD and CW is starting to take on.
\citet{DBLP:conf/clef/SawinskiWKSLSA23} and \citet{hyben2023bigger} compare the performance of fine-tuned language models such as BERT with LLMs using zero- and few-shot learning as well as fine-tuning. Although zero- and few-shot approaches for LLMs underperform, the authors note their reliance on internal definitions of worthiness and limited prompt testing. As part of the fully automated fact-checking system relying only on LLMs, \citet{li2023selfchecker} implement a CD module using a verbose few-shot prompt, yet they do not report performance metrics.
Finally, \citet{ni2024afacta} tackle CD by proposing a three-step prompting approach to examine model consistency. However, neither 
\citet{li2023selfchecker} nor \citet{ni2024afacta} address the CW task. To our knowledge, there is no work on CW focused on describing specific worthiness criteria using verbose prompts.

% fine-tuning LLMs \citep{DBLP:conf/clef/AgrestiHC22} 
% data augmentation with LLMs in combination with PLMs\citep{DBLP:conf/clef/ModzelewskiSW23}. 
%In their limitations they state both their disregard for worthiness detection as well as not attempting to use annotation guidelines in prompts. However, solving check-worthiness with LLMs is also worth exploring, despite the difficulty of pinpointing the worthiness criterion and its inherent subjectivity.

\section{Datasets }

\paratopic{Presenting the datasets}

Our experiments utilize five datasets in English covering diverse topics and genres. 
Examples from each dataset are presented in Table~\ref{tab:examples}. We next describe each dataset in more detail, including the CD and CW criteria used.

\begin{table*}
\centering
 \begin{adjustbox}{width=\textwidth}
    {
    \small
        \begin{tabular}{@{}lcl@{}}
            \toprule
            \textbf{Dataset}  & \textbf{Label} & \textbf{Example}  \\ 
            \midrule
            \multirow{3}{*}{\textbf{CB}} & \xmark & \sethlcolor{lightred}\hl{\textit{I would do the opposite in every respect.}}\\
            & \textit{\textbf{O}} & \sethlcolor{lightyellow}\hl{\textit{I have met with the heads of government bilaterally as well as multilaterally.}} \\
            & \cmark & \sethlcolor{palegreen}\hl{\textit{Fifty percent of small business income taxes are paid by small businesses.}} \\
            \midrule
            \multirow{4}{*}{\textbf{CLEF}} & \xmark & \sethlcolor{lightred}\hl{\textit{If the vaccine was dangerous they would've given it to poor people first, not politicians and billionaires.}} \\
            & \multirow{2}{*}{\textit{\textbf{O}}} & \sethlcolor{lightyellow}\hl{\textit{Today, FDA approved the first COVID-19 vaccine for the prevention of \#COVID19 disease in individuals}} \\
            & & \sethlcolor{lightyellow}\hl{\textit{16 years of age and older.}} \\
            & \cmark & \sethlcolor{palegreen}\hl{\textit{They said the vaccine stopped transmission. Now they are lying and saying they didn’t. Video proof here}}\\
            \midrule
            \multirow{2}{*}{\textbf{ENV}} & \xmark & \sethlcolor{lightred}\hl{\textit{We Love Green! The environment is at the heart of Parisian electro-pop music festival We Love Green.}} \\
            & \cmark & \sethlcolor{palegreen}\hl{\textit{All pension fund clients have a target for carbon reduction of the equity investments.}} \\
            \midrule
            \multirow{4}{*}{\textbf{NEWS}} & \multirow{2}{*}{\xmark} & \sethlcolor{lightred}\hl{\textit{In Germany, RT has also amplified voices questioning the threat of COVID-19, and calling testing }} \\
            & &  \sethlcolor{lightred}\hl{\textit{and mask-wearing into question.}} \\
            & \multirow{2}{*}{\cmark} & \sethlcolor{palegreen}\hl{\textit{"If you wash and dry a cloth face mask on high heat, then you should be good to go," according to }}\\
            & & \sethlcolor{palegreen}\hl{\textit{professor Travis Glenn.}} \\
            \midrule
            \multirow{2}{*}{\textbf{POLI}} & \xmark & \sethlcolor{lightred}\hl{\textit{As I have said all along, the courts are where we will win this battle.}}\\
            & \textit{\textbf{O}} & \sethlcolor{palegreen}\hl{\textit{I promised that our roads would be the envy of the nation.}}\\
            \bottomrule
        \end{tabular}}
        \end{adjustbox}
\caption{Examples from the datasets used. \xmark = non-factual claim, \emph{\textbf{O}} = factual claim, \cmark = check-worthy claim }
\label{tab:examples}
\end{table*}

%
%Been emphasizing the emerging picture of brain injury from COVID19 + longCovid. Antivaccine forces keep pointing to low death rates in young people and adolescents, while ignoring the rapidly accelerating hospitalizations and now neurological complications

%\begin{description}
\paragraph{ClaimBuster (CB)}\citep{hassan2017toward} is a widely used dataset of claims from USA presidential debates, featuring ternary labels (\emph{non-factual}, \emph{unimportant factual}, \emph{check-worthy factual}) that distinguish between check-worthy and unimportant factual claims. This setup addresses both the CD and CW tasks. Claims are deemed check-worthy if the general public would be interested in their veracity. However, no specific definition of factuality is provided -- unimportant factual claims are defined as those lacking check-worthiness.

\paragraph{CLEF CheckThat!Lab 2022 (CLEF)}\citep{alam-etal-2021-fighting-covid} contains tweets about COVID-19, with two parts: a set of tweets with claims and a subset with check-worthy claims, addressing both CD and CW tasks. Check-worthiness is defined as the need for professional fact-checking, excluding jokes, trivial claims, or those deemed uninteresting. Factual claims are defined as sentences that assert something is true and can be verified using factual information, such as statistical data, specific examples, or personal testimony.

\paragraph{EnvironmentalClaims (ENV)}\citep{stammbach-etal-2023-environmental} is compiled from environmental articles and reports. The dataset focuses on check-worthy environmental claims related to green-washing in marketing strategies. 
The authors defined specific criteria for an environmental claim that extend beyond the topic itself (e.g., highlighting the positive environmental impact of a product, not being too technical). The annotators were instructed to label only the explicit claims, discouraging the selection of claims with inter-sentence coreferences.

\paragraph{NewsClaims (NEWS)} \citep{gangi-reddy-etal-2022-newsclaims} comprises sentences from news articles on COVID-19, with metadata available for positives (speaker, object, claim span). The annotators were asked to judge whether a claim falls into one of the four topic-specific categories, which essentially formed the worthiness criteria, even though check-worthiness was not explicitly mentioned in the guidelines. The dataset includes both check-worthy and non-check-worthy claims with inter-sentence coreferences (e.g., \emph{That's also false}), which typically require inspecting the surrounding context to determine their check-worthiness (we estimate this applies to about 10\% of claims in the test set).
% \emph{That's also false.}
% \emph{There is no evidence to back up this rumor, which has circulated on the internet.}
% \emph{Is using UV light to kill Covid-19 effective?}
% \emph{It was initially linked to a live animal market in the city of Wuhan, in the Hubei province of China.}

\paragraph{PoliClaim (POLI)}\citep{ni2024afacta} covers the same topic as ClaimBuster (politics, speeches of governors) but labels only verifiable claims, leaving out check-worthiness. The authors provided detailed guidelines on verifiable claims, emphasizing the need for specificity and differentiation from opinions lacking factual basis.
To handle ambiguous cases, they employed a ternary (\emph{Yes}, \emph{No}, \emph{Maybe}) annotation scheme. \emph{Maybe} indicates that a claim may contain factual information but does not fully meet all criteria. For claims labeled as \emph{Maybe}, annotators answered a follow-up Yes-No question to determine whether the claim leans toward factual information or subjective opinion. As with NEWS, inter-sentence coreference was considered; since the claims are extracted from political speeches, many of them include personal pronouns (\emph{I}, \emph{we}), which necessitates coreference resolution to identify the claimant or subject.

%\end{description}

\begin{table*}[t]
\centering
    {
    \small
        \begin{tabular}{lccccc}
            \toprule
             & \textbf{CB} & \textbf{CLEF} & \textbf{ENV} & \textbf{NEWS} & \textbf{POLI}  \\ 
            \midrule
            Task & CD+CW & CD+CW & CW & CW & CD\\ 
            Labels & ternary & binary* & binary & binary & binary* \\
            \# instances & 23,533 & 3,040 & 2,647 & 7,848 & 52 speeches\\ 
            \# instances used & 1,032 & 251  & 275 & 1,622 & 816 \\
            label distribution & 731/238/63 & 102/110/39 & 198/67 & 811/811 & 295/521 \\
            Genre & debates & tweets & news articles & reports & speech transcripts \\ 
            Topic & politics & healthcare & environment & healthcare & political \\
            Co-text & 4 preceding, on request & -- & not available & inconclusive & 1 preceding, 1 following \\
            Agreement & --* & 0.75/0.7 & 0.47 & 0.405 & 0.69 \\ 
            Agreement metric & -- & Fleiss-$\kappa$ & Krippendorff-$\alpha$ & Krippendorff-$\kappa$ & Cohen-$\kappa$\\
            \bottomrule
        \end{tabular}}
\caption{Characteristics of the CD and CW datasets used in our experiments. *CB reported no agreement evaluation, but the test set used is agreed upon by experts. Label distribution in order: \xmark / \emph{\textbf{O}} / \cmark}
\label{tab:datasets}
\end{table*}

\medskip
\paratopic{Things I want to say about the datasets}

We use these five datasets because they provide detailed annotation guidelines and cover various topics, genres, and worthiness criteria. Table~\ref{tab:datasets} summarizes their characteristics (see Appendix~\ref{sec:dataset_info} for details). The CB and CLEF datasets address both CD and CW tasks, with CB using ternary labels annotated together and CLEF using binary labels with separate questions for CD and CW.
%ClaimBuster and CLEF cover both tasks, PoliClaim only CD, and NewsClaims and EnvironmentalClaims only CWD. 
The five datasets were originally annotated using a binary scheme (ENV), Likert scale (CLEF-CW), multi-class (NEWS), or a follow-up prompt for uncertain instances (POLI).
All datasets have aggregated binary labels, except CB, where aggregation from ternary into binary labels is straightforward.
The reported inter-annotator agreement is substantial for POLI and CLEF \citep{Landis1977}, but moderate for ENV and NEWS, reflecting the complexity of the domain-dependent CW task.

%that the effort to reduce vagueness with highly detailed annotation guidelines is sometimes not enough for a task like this -- subjectivity?
%The distribution of labels was given by some datasets, which allows for the detection of challenging instances and can be used for further analysis of the models' performance.

\section{Experimental Setup}

In our experiments, we use both closed-source and open-source LLMs. For closed source, we use OpenAI models \emph{gpt-turbo-3.5} and \emph{gpt-4-turbo}. For open-source models, due to hardware constraints, we chose Llama 3 8B Instruct, which is the top performer in its parameter class. To ensure reproducibility and encourage deterministic behaviour, we prompt GPT models with the temperature setting of 0 along with a fixed seed parameter and use greedy sampling with top\_p=1 for open-source models.
We also experimented with other open-source models. Mistral 7B Instruct v0.2 was not compliant with the provided labels, instead giving open-ended answers, even for less verbose prompts. 
See Appendix~\ref{sec:models} for more detailed information on models. 

\subsection{Prompt Verbosity}

We first investigate how prompt verbosity affects LLMs' predictive accuracy. We hypothesize that the optimal verbosity level depends on the dataset, reflecting the factuality and worthiness criteria differences between the domains. While a brief prompt might lack essential details, a comprehensive prompt featuring extensive definitions and examples may make the task more difficult to solve. Across datasets and for each prompt level, we aim to preserve the original wording and typography of the annotation guidelines as much as possible since we aim to establish whether guidelines without much intervention can be used as prompts for up-to-par performance. We additionally instruct the model to reply using only the provided labels without additional explanation to increase compliance and streamline evaluation. For POLI, we use the same question structure as in the annotation -- for instances where the model responded with \emph{Maybe}, we prompt it again with the follow-up question, providing previous responses in the prompt.

Based on the content and style of annotation guidelines, we define the following four levels of verbosity (cf.~Appendix~\ref{sec:prompts} for full prompts for four verbosity levels across the five datasets):

\begin{description}

%\paragraph{Level V0} 
\item[Level V0]
serves as the baseline. We use a naive zero-shot prompt, relying on internal definitions of the model. For the CD task (for the CB, CLEF and POLI datasets), we use the following prompt: 
%
%\begin{quote}
``\emph{Does the following sentence/statement/tweet contain a factual claim? Answer only with Yes or No.}''
%\end{quote}
%
For the CW task (for the CB, CLEF, NEWS and ENV datasets) we use the following prompt:
%
%\begin{quote}
``\emph{Does the following sentence/statement/tweet contain a check-worthy claim? Answer only with Yes or No.}''
%\end{quote}
%
As these prompts do not include the specific factuality or worthiness criteria from the guidelines, they serve as a domain-agnostic baseline;
    
%\paragraph{Level V1}
\item[Level V1]
uses prompts that include the task definition and the set of possible labels but omit detailed explanations of the labels or principles. For example, for the CB dataset, the three categories of non-factual, unimportant factual, and check-worthy factual sentences are introduced but not explained;

%\paragraph{Level V2} 
\item[Level V2]
expands on V1 by adding a more detailed explanation of the labels or general annotation principles (or both, in the case of PoliClaim). Some principles include avoiding implicit assumptions (ENV), defining check-worthiness criteria based on public interest (CB), and categorizing claims that non-professionals can verify as non-check-worthy (CLEF);

%\paragraph{Level V3} 
\item[Level V3]
builds on V2 by including examples from the original annotation guidelines. This level closely aligns with annotation guidelines, encompassing all or nearly all information the datasets' authors provide in their accompanying papers.\footnote{CB and ENV documented additional examples (typically 20--30 examples) provided to the annotators. We did not include these examples.} The examples are provided either along with the labels (CB), separately in a few-shot fashion (ENV), or both (POLI).
%This level is closest to annotation guidelines, encompassing all or most information that the datasets' authors provided in the papers accompanying the datasets. 
%\end{description}

\end{description}

% \medskip

\subsection{Amount of Context}

%\citet{ni2024afacta} argue co-reference needs to be resolved for determining whether a claim is verifiable, while \citet{gangi-reddy-etal-2022-newsclaims} advocate the importance of the speaker, claim object, claim span and other metadata was disregarded.
In real-world scenarios, claims are rarely evaluated in isolation. Accordingly, annotators working with CD and CW datasets were usually provided with some contextual information, consisting of the claim's co-text and metadata. Regarding co-text, the quantity varied between datasets (cf. Table~\ref{tab:datasets}), as did its significance -- sometimes it was provided as additional guidance (CB), while in other cases, it was deemed crucial for assigning labels (POLI, NEWS).
For NEWS, the amount of provided co-text is inconclusive, so we decided to omit it from co-text expansion.
This difference highlights that co-text is both another undefined aspect of CD and CW, and that it can vary across domains. Similarly, metadata such as speaker, affiliation, occasion, and date were revealed only during annotation for CLEF-CW and were not available in the dataset itself. However, metadata is available for the CB and POLI datasets, while NEWS provides metadata only for positives, making it unusable for our experiments. Adding metadata might lead to biases, yet it could offer essential information, depending on the worthiness criterion. 

%Claims are never made in isolation; their context matters not only for verifying their veracity but also for gauging their factuality and worthiness even before fact-checking.
We investigate how LLMs' predictive accuracy depends on the amount of situational context provided to the model. To this end, we leverage the context information available in the CB and POLI datasets and expand the prompts in three variants:

\begin{table*}[t!]
\centering
    {
    \small
        \begin{tabular}{lcccccccc}
            \toprule
             && \multicolumn{2}{c}{\textbf{CB}} & \multicolumn{2}{c}{\textbf{CLEF}} & \textbf{ENV} & \textbf{NEWS} & \textbf{POLI} \\ 
            \cmidrule(lr){3-4}\cmidrule(lr){5-6}\cmidrule(lr){7-7}\cmidrule(lr){8-8} \cmidrule(lr){9-9}
            & & CD & CW & CD & CW & CW & CW & CD \\ 
            \midrule
            Stratified random & & .375 & .452 & .745 & .415 & .25 & .667 & .779 \\
            % TF-IDF SVM & & & & & & & & \\
            Previous & & .818$^*$ & .818$^*$ & .761$^{a}$ & .698 & .849 & .309$^*$ & .862$^{a}$ \\
            SVM && .789 & .799 & .726 & .346 & .729 & .675$^*$ & .819\\
            BERT && .956 & .938 & .773 & .472 & .822 & .771$^*$ & .881 \\
            \midrule
            \midrule
            \multirow{4}{*}{gpt-4} & V0 & .833 & .805 & .797 & .467 & .416 & \textbf{.583} & \textbf{.844}\\ 
            & V1 & .883 & .885 & .799 & .552 & \textbf{.773} & .572 & .679 \\ 
            & V2 & .908 & .889 &\textbf{ .806} & \textbf{.583} & .690 & .480 & .541 \\ 
            & V3 & \textbf{.919} & \textbf{.927} & .781 & .556 & .596 & .523 & .563\\
            \midrule
            \multirow{4}{*}{gpt-3.5} & V0 & .853 & .718 & .656 & \textbf{.496}  & .484 & \textbf{.531} &.707 \\ 
            & V1 & .570 & .739 & .490 & .438 & \textbf{.710} & .371 & .751 \\ 
            & V2 & .774 & .800 & .650 & .468 & .701 & .348 &.657 \\ 
            & V3 & \textbf{.872} & \textbf{.862} & \textbf{.757} & .446 & .650 & .206 &\textbf{.803} \\
            \midrule
             \multirow{4}{*}{Llama3 8B} & V0 & .677 & .743 & .769 & \textbf{.439} & .290  & \textbf{.586}  & .812\\ 
            & V1 & .478 & .655 & .803 & .415 & \textbf{.755} & .502& \textbf{.827} \\ 
            & V2 & \textbf{.742} & \textbf{.751} & \textbf{.807} & .433 & .745 & .466 &.712 \\ 
            & V3 & .702 & .637 & .790 & .426 & .742 & .469 & .651 \\
            \bottomrule
        \end{tabular}} 
\caption{Binary F1 scores across datasets and prompt verbosity levels (V1--V3). Level V0 corresponds to the naive-prompting baseline. For baselines and previous results: $^{a}$ = accuracy, $^*$ = not directly comparable}
\label{tab:levels_detailed}
\end{table*}

\begin{description}
    
%\paragraph{Level C1} 
\item[Level C1]
represents adding the co-text of the claim. The amount of co-text included in the prompt for each dataset is the same as what was originally shown to the annotators -- for CB, four preceding statements (which were either by the speaker, opposing speaker, or moderator), and for POLI, one preceding and one following statement;

%\paragraph{Level C2} 
\item[Level C2]
expands the contextual information by adding metadata to the claim. In the case of POLI, the metadata is the speaker's identity and political party, whereas for CB it additionally contains the speaker's title and the sentiment of the statement, provided by the authors of the dataset;

%\paragraph{Level C3} 
\item[Level C3]
combines both C1 and C2 by providing both co-text and metadata.

\end{description}

We appended the contextual information to the user prompts, and only modified the system prompts of POLI slightly -- adding guidance on how to handle context, ommited from the no-context variants (cf.~Appendix~\ref{sec:dataset_info} for a detailed description).

\section{Results}

%\subsection{Baselines}

We present the results for prompt verbosity levels in Table~\ref{tab:levels_detailed} and for different context levels in Table~\ref{tab:levels_context}.
In Table~\ref{tab:levels_detailed}, we also include the previous results reported by authors in the original papers introducing the datasets (note that some results are not directly comparable to ours, as we discuss below). \alert{We use a stratified random classifier, an SVM classifier with TF-IDF features, and a fine-tuned BERT \citep{devlin2019bert} as baselines.}

\subsection{Prompt Verbosity}

Table~\ref{tab:levels_detailed} shows the baselines and F1 scores by verbosity level for \emph{gpt-4-turbo}, \emph{gpt-3.5-turbo}, and Llama3 8B. Both performance and the optimal verbosity level is not consistent across datasets. The accuracy generally increases with verbosity levels for CB, but the trend is reversed for ENV. We observe no consistent trend for CLEF, POLI, and NEWS datasets. The most verbose prompts (V3) generally do not achieve the highest performance, except for the GPT models and CB. This highlights that providing detailed instructions and examples can be beneficial but potentially harm performance.

\paragraph{Comparison to baselines.}
For SVM and BERT baselines, we had to use a different test set for NewsClaims than for the other models, as the original dataset does not provide a training set (the best-performing LLM on this test set achieved an F1 score of 0.670). Overall, all best-performing LLMs outperform the SVM baseline. However, except for CLEF, BERT outperforms the best-performing LLMs by a small margin (<0.05 F1), raising doubts about whether manual data labeling is worthwhile in these cases.

\paragraph{Comparison to previous work.} 
\alert{Comparing with previous work is difficult due to differences in setup.} For CB, the authors evaluated used 4-fold cross-validation on different-sized subsets (4,000, 8,000 \dots 20,000), all containing our chosen test set, annotated by experts. The authors evaluated using weighted F1-score, achieving a maximum score of 0.818. Our highest weighted F1-scores surpass this, reaching 0.933 for \emph{gpt-4-turbo} and 0.906 for \emph{gpt-3.5-turbo}.
On CLEF, the best-reported result is the accuracy score of .761 for CD and the F1 score of .698 for CW. While our approach underperforms for CW (F1 of 0.583), it achieves higher accuracy for CD (0.776 on Level V2).
In the case of NEWS, the authors reported an F1 score, but it remains unclear whether it was evaluated based on binary or multiclass labels, given that annotators had to categorize claims into different classes. They achieved the highest F1 score of 0.309, which our approach exceeds on the subset we selected, achieving an F1 score of 0.583. Our subset has a higher random baseline due to a higher ratio of positive examples and includes all positives from the original test set. 
For POLI, the authors evaluated using accuracy. They achieved an accuracy of 0.764 on the test set using \emph{gpt-3.5} and 0.862 using \emph{gpt-4}. The \emph{GPT-3.5-turbo} using prompt Level V3 performs comparable, while \emph{GPT-4} and Llama3 perform worse.
For ENV, the metrics are directly comparable, and our approach underperforms compared to previous results.

\begin{table}
\centering
    {
    \small
        \begin{tabular}{@{}lccccccc@{}}
            \toprule
             & \multicolumn{2}{c}{\textbf{CB}} & \multicolumn{2}{c}{\textbf{CLEF}} & \textbf{ENV} & \textbf{NEWS} & \textbf{POLI} \\ 
            \cmidrule(lr){2-3}\cmidrule(lr){4-5}\cmidrule(lr){6-6}\cmidrule(lr){7-7} \cmidrule(lr){8-8}
            & CD & CW & CD & CW & CW & CW & CD \\ 
            \midrule
            V1 & 26 & 9  & 8  & 55 & 13 & 150 & 87 \\
            V2 & 4  & 3  & 5  & 35 & 8  & 90  & 87 \\
            V3 & 1  & 1  & 2  & 14 & 5  & 68  & 65 \\
            \bottomrule
        \end{tabular}} 
        \caption{Counts of instances misclassified by all three models across the three verbosity levels.}
\label{tab:errors_levels}
\end{table}

\begin{table*}
\centering
 \begin{adjustbox}{width=\textwidth}
    {
    \small
        \begin{tabular}{@{}lccl@{}}
            \toprule
            \textbf{Dataset}  & \textbf{Label} & \textbf{Gold} & \textbf{Example}  \\ 
            \midrule
            \multirow{1}{*}{\textbf{CB}} & 1 & 0 & \textit{Let's go to work and end this fiasco in Central America, a failed policy which has actually increased Cuban and Soviet influence.} \\
            \midrule
            \multirow{2}{*}{\textbf{CLEF}} & \multirow{2}{*}{1} &  \multirow{2}{*}{0} & \textit{So businesses will get fined \$14,000 (per employee) if they don't comply with Biden's vaccine mandate. And illegal aliens} \\
            &  &  & \textit{get a \$450,000 payout for "damages" for crossing our border illegally... Biden's America.} \\
            \midrule
            \multirow{2}{*}{\textbf{ENV}} & 1 & 0 & \textit{Renewable energy is purely domestic sourced and environment-friendly, and can be used continuously without being depleted.} \\
            & 0 & 1 & \textit{To strengthen its approach, Kering’s SBT for a 1.5°C trajectory was revised and approved by the SBTi in early 2021.} \\
            \midrule
            \multirow{2}{*}{\textbf{NEWS}} & 1 & 0 & \textit{There’s currently no strong evidence that supplementing with vitamin C will prevent or cure COVID-19.} \\
            & 0 & 1 & \textit{Vaccines, by their nature, are reactive.} \\
            \midrule
            \multirow{2}{*}{\textbf{POLI}} & 0 & 1 & \textit{We are finally going to fix the darn roads.} \\
            & 0 & 1 & \textit{Too many people are struggling to make ends meet.} \\
            \bottomrule
        \end{tabular}}
        \end{adjustbox}
\caption{Examples of characteristic instances per dataset that were consistently misclassified across levels.}
\label{tab:error_examples}
\end{table*}

\begin{table*}[t!]
\centering
    {
    \small
        \begin{tabular}{@{}lcccccccccc@{}}
            \toprule
             & &  \multicolumn{6}{c}{\textbf{CB}}  & \multicolumn{3}{c}{\textbf{POLI}} \\
            \cmidrule(lr){3-8} \cmidrule(lr){9-11}
            & & \multicolumn{3}{c}{CD} & \multicolumn{3}{c}{CW} & \multicolumn{3}{c}{CD} \\
            \cmidrule(lr){3-5} \cmidrule(lr){6-8} \cmidrule(lr){9-11}
            & & V1 & V2 & V3 & V1 & V2 & V3 & V1 & V2 & V3 \\
            \midrule
            \multirow{4}{*}{gpt-4-turbo} & C0 & \textbf{.883} & \textbf{.908} & \sethlcolor{palegreen}\hl{\textbf{.919}} & \textbf{.885} & .889 & \sethlcolor{palegreen}\hl{\textbf{.927}} & .619 & .541 & .563 \\ 
            & C1 & .806 & .849 & .862 & .803 & .847 & .872 & .722 & \textbf{.650} & .727  \\ 
            & C2 & .879 & \textbf{.908} & .913 & .880 & \textbf{.901} & .916 & \textbf{.707} & .470 & .592\\
            & C3 & .794 & .857 & .877 & .791 & .854 & .885 & .692 & .632 & \sethlcolor{palegreen}\hl{\textbf{.732}} \\
            \midrule
            \multirow{4}{*}{gpt-3.5-turbo} & C0 & \textbf{.570} & .774 & \sethlcolor{palegreen}\hl{\textbf{.872}} & .739 & .800 & \sethlcolor{palegreen}\hl{\textbf{.862}} & .751 & .657 & \sethlcolor{palegreen}\hl{\textbf{.803}} \\ 
            & C1 & .461 & .299 & .513 & .517 & .301 & .528 & .790 & .688 & .794 \\ 
            & C2 & .560 & \textbf{.801} & .836 & \textbf{.747} & \textbf{.826} & .832 & .730 & .523 & .704 \\
            & C3 & .474 & .724 & .758 & .643 & .716 & .749 & \textbf{.794} & \textbf{.754} & .800 \\
            \midrule
            \multirow{4}{*}{Llama3 8B} & C0 & .478 & .742 & .702 & .655 & .751 & .637 & \sethlcolor{palegreen}\hl{\textbf{.827}} & .712 & .651  \\ 
            & C1 & .460 & .591 & .614 & .531 & .528 & .552 & .799 & .789 & .803  \\ 
            & C2 & \textbf{.483} & \sethlcolor{palegreen}\hl{\textbf{.773}} & \textbf{.764} & \textbf{.727} & \sethlcolor{palegreen}\hl{\textbf{.819}} & \textbf{.736} & .807 & .703 & .628 \\
            & C3 & .468 & .610 & .601 & .506 & .618 & .556 & .806 & \textbf{.798} & \textbf{.805} \\
            \bottomrule
        \end{tabular}}
\caption{Binary F1 scores by level of context information (C1--C3) added to the prompt ranging in verbosity (V1--V3) . Level C0 corresponds to the prompt level with no context information. The best scores across verbosity levels are shown in bold, and the best scores per model and dataset are highlighted in green.}
\label{tab:levels_context}
\end{table*}

\paragraph{CD vs.~CW.} Generally, higher performance is achieved for the CD task, although the diverse domains of the datasets and differences in guidelines prevent definitive conclusions. Therefore, comparing performance on the two datasets that cover both tasks -- CB and CLEF -- is most straightforward. Interestingly, a reverse phenomenon is observed between these datasets—significantly higher performance is achieved for the CD task on CLEF, whereas on CB, CW performance is slightly higher.
An important difference in the two datasets is precisely in the annotation styles -- CB uses the same guidelines for both tasks and ternary annotation, while for CLEF the guidelines are different for the two tasks, originally using different labelling strategies (binary for CD and Likert scale for CW). 

\paragraph{Closed-source vs.~open-source.} While broader conclusions require a wider range of both open- and closed-source models, especially larger open-source ones, the Llama3 8B model performs similarly to GPT models, highlighting the potential of prompting open-source models with annotation guidelines. Furthermore, the results of both GPT models on the CB dataset could indicate a potential data leakage \cite{balloccu-etal-2024-leak} since the performance of Llama3 8B is comparable in other datasets but lags for CB.

\subsection{Error Analysis}

\paragraph{Worst performance.} The naive baseline prompt (V0) generally outperforms the prompts based on annotation guidelines on the CLEF CW and NEWS datasets, except for V2 for CLEF CW with \emph{gpt-4-turbo}. 
For CLEF CW, the annotation guidelines are adapted from the Likert scale, where multiple characteristics are attributed to negatives (e.g., not interesting, a joke, not containing claims, or too trivial to be checked by a professional). In our prompts, we converted the Likert scale to binary, where the already diverse and vaguely defined criteria were binned in a single label, increasing complexity.
For NEWS, although the dataset's purpose is claim check-worthiness detection, check-worthiness as a concept is not mentioned in the annotation guidelines. Positives are merely selected by containing claims falling into four predefined categories relating to the COVID-19 virus, and check-worthiness is assumed implicitly. This, along with the presence of inter-sentence coreference in the positive instances, might cause poor performance.

\paragraph{Most difficult instances.}
To analyze poor performance beyond the F1 score, we decided to identify the instances for which all three models across levels consistently predicted the wrong label. Table~\ref{tab:errors_levels} shows the counts of those instances.

Interestingly, whether the instances in question were consistently misclassified as positives or negatives depends on the domain -- for CLEF and CB, all of the instances are false positives (FP), whereas for POLI, all 65 are false negatives (FN). 
This suggests that the guidelines are too restrictive regarding the positive label or are interpreted as such during inference.
%For ENV and NEWS the final set contains both FP and FN, yet certain patterns can be observed which distinguish .

Table~\ref{tab:error_examples} shows examples from the final pool of mislabelled instances. Several interesting observations can be made here. For example, CLEF comprises tweets, where sarcasm is more prevalent, making the prediction task harder. For ENV, which contains environmental reports requiring expert knowledge, the mislabeled instances were either too vague for positives or contained too much domain knowledge for the models to decipher. Annotators were urged to look up acronyms they were not familiar with, but the same could not be accomplished with ICL. For NEWS, some claims are only implicitly related to COVID-19, which results in a false negative label. On the other hand, some instances seem mislabeled in the gold set. Concerning POLI, the frequent use of personal pronouns requiring coreference resolution leads to false negatives, with the claim losing relevance without information about the claimant.

\subsection{Amount of Context}

Table~\ref{tab:levels_context} shows the F1 scores by verbosity and context level for all models.
The benefit of including context varies across models -- there is a bigger performance increase for the Llama model with added contextual information, topping the performance for CB in both tasks as opposed to prompts with no context. For the GPT models, there is some positive impact of metadata (C2).
The least beneficial is the addition of co-text \alert{but no metadata} (C1), including speaker information, which is vital when given previous responses.  Concerning prompt verbosity levels, context's impact is higher on less verbose prompts, showing contextual information complements brief definitions.

%This could be due to lack of clarity caused surrounding sentences or bias caused by speaker information.

\subsection{Rank-Based Evaluation}

\begin{figure*}[t]
\centering
% \hspace*{-0.8cm} 
\includegraphics[width=\textwidth]{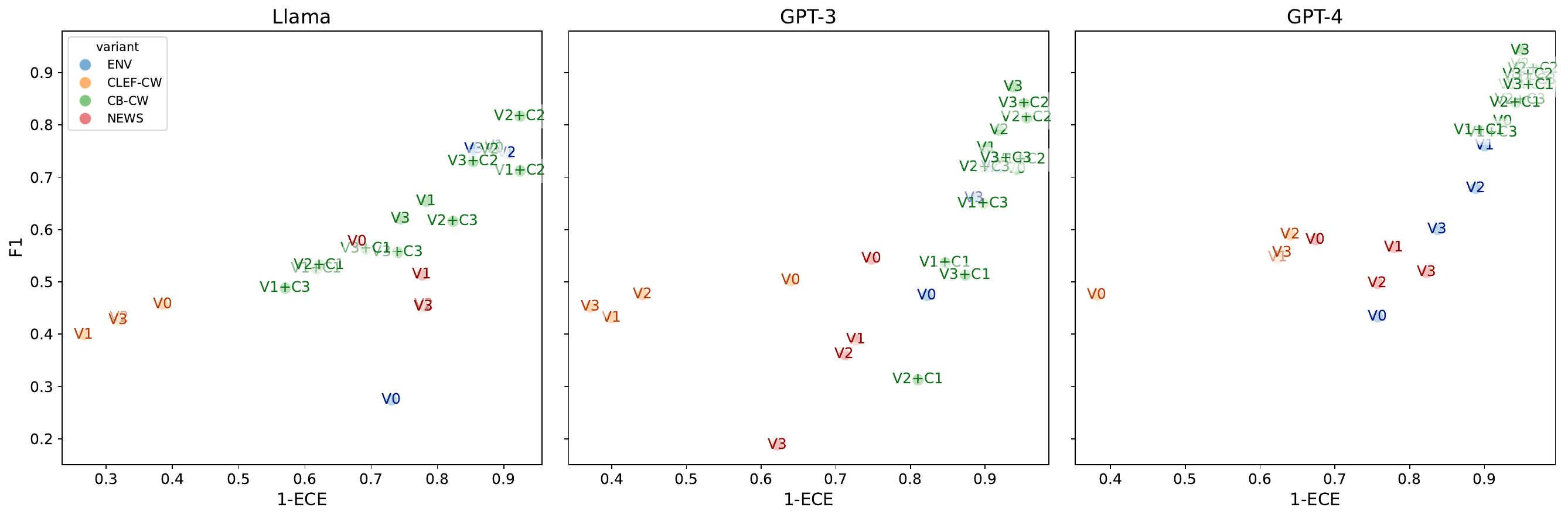}
\caption{F1 scores and calibration accuracy ($1-\mathrm{ECE}$) for the CW task, across datasets and models}
\label{fig:f1-calibration}
\end{figure*}

In light of resource constraints, fact-checking organizations have devised principles to prioritize claims based on their check-worthiness. This invites the question of whether zero- and few-shot LLM prompting could be used for that purpose. To investigate this, we frame CW as a \alert{binary relevance ranking} and rank the claims based on the LLM's confidence for the positive class. We used the token likelihood of the positive class as a measure of confidence.
The quality of the so-obtained ranking will depend on how well the LLM is calibrated. Thus, we first evaluate the LLMs' calibration accuracy using the expected calibration error (ECE). Figure~\ref{fig:f1-calibration} shows the predictive accuracy (F1 score) against calibration accuracy ($1-\mathrm{ECE}$) across datasets and prompt verbosity levels (we only use prompts at context level C0, i.e., no context information). 

\begin{table}[t]
\centering
    {
    \small
        \begin{tabular}{llrrrr}
            \toprule
             & & \textbf{CB} &  \textbf{CLEF}  & \textbf{ENV} & \textbf{NEWS} \\
            \midrule
            \multirow{3}{*}{gpt-4} & AP & .951  & .552 & .767 & .67   \\ 
            & P\@10 & 1   & .9   & .9   & 1   \\
            & P\@R & .924 & .615 & .761 & 1   \\
            \midrule
            \multirow{3}{*}{gpt-3.5} & AP & .934  & .464 & .796 & .669  \\ 
            & P\@10 & 1   & .6   & .9  &  .7  \\
            & P\@R & .919 & .436 & .772 & .700 \\
            \midrule
            \multirow{3}{*}{Llama3 8B} & AP & .878  & .350 & .794  &  .688  \\ 
            & P\@10 & 1   & .2   &  .1  &  1  \\
            & P\@R & .823 & .282 &  .762 &  1  \\
            \bottomrule
        \end{tabular}}
\caption{Rank-based CW performance scores}
%for the CW task: average precision (\emph{AP}), precision-at-10 (\emph{P@10}), and precision-at-R (\emph{P@R})}
\label{tab:ranking}
\end{table}

Per model and dataset, we select the prompt that scores high on predictive and calibration accuracy. The prompts with the highest F1 scores are usually the best-calibrated ones, except for NEWS, where we select level V1 as Pareto-optimal.

Table~\ref{tab:ranking} shows the rank-based performance scores for the selected prompts: average precision (AP), precision-at-10 (P@10), and precision-at-R, where R equals the total number of positives in the dataset. The rank-based performance scores mirror the classification accuracy scores: they are high for datasets with high predictive accuracy (CB and ENV) and lower for datasets with lower predictive accuracy (NEWS and CLEF). Our results suggest that LLM models with high predictive accuracy also produce well-calibrated scores using ECE and may be readily used as check-worthiness rankers.

\section{Conclusion}

We tackled claim detection and check-worthiness tasks using zero- and few-shot LLM prompting based on existing annotation guidelines. 
The optimal level of prompt verbosity, from minimal prompts to detailed prompts that include criteria and examples, varies depending on the domain and guidelines style. Adding claim context %(co-text and speaker information) 
does not improve performance. Models with high predictive accuracy can directly utilize confidence scores to produce reliable check-worthiness rankings.

\section*{Limitations}

\paragraph{Datasets.} Our experiments do not use datasets created by fact-checking organizations. While the datasets were created specifically for the tasks of CD and CW, and most were annotated by experts, the datasets were constructed for research purposes. To most accurately evaluate the potential of using our approach in fact-checking organizations, a dataset annotated according to official factuality or check-worthiness criteria with appropriate annotation guidelines should be used.

\paragraph{Models.} Due to hardware constraints, no open-source LLMs greater than 8B parameters were used in our experiments. We acknowledge the importance of relying on open-source models in the research community and the lack of insight that results from disregarding larger open-source models. Using closed-source models has the additional caveat of possible leakage of the dataset, which is a growing concern in the community \citep{balloccu-etal-2024-leak}. We also note that the outstanding results on the ClaimBuster dataset (CB) could be due to data leakage, considering the dataset was published several years ago and has a wide reach in the research of automatic fact-checking. 

\paragraph{Languages.} In this work, we only do experiments on datasets in English. This is for two reasons: (1) the necessity to understand the annotation guidelines to draft prompts using them and (2) the lack of datasets in other languages. However, we acknowledge that disinformation is a global problem and that tackling it requires working with multiple languages.

\paragraph{Lack of prompt engineering experiments.}
In this work, we do minimal prompt engineering interventions beyond merely adapting the level of detail in annotation guidelines and appending contextual information. We opted for this approach instead of drafting prompts ourselves to investigate how original wording, definitions, and examples given to annotators could fare with LLMs. We realize weak performance in some cases (e.g., CLEF, for the naive aggregation from the Likert scale to binary labels), and performance variations could be due to the models' sensitivity to prompt structure, wording, and examples. However, translating the complex criteria of worthiness in such a streamlined way could benefit fact-checkers. Furthermore, prompt design should be adapted for each dataset, significantly expanding the scope of this research (since five datasets are used).
We leave experiments regarding prompt design for future work.

\alert{%
\paragraph{Binary relevance ranking.}
In our study, we also assess CW as a ranking task, which is crucial given that fact-checking organizations often face time and resource constraints and must prioritize claims based on their CW criteria. We use binary relevance judgments for evaluation, with binary CW labels as the ground truth, and apply standard IR metrics such as MAP and P@R. An alternative approach could involve using graded CW criteria, framing the task as regression, and evaluating with graded relevance judgments like NDCG. However, to our knowledge, only the dataset by \citet{gencheva-etal-2017-context} provides graded CW labels, using aggregated judgments from various fact-checking agencies to model priority.}

% \paragraph{CLEF\&NEWS.} The results of weak performance on the CLEF and NEWS datasets could lie in the worthiness criteria used and the way the criteria are articulated. In CLEF, a Likert scale was used for the annotation. However, the levels in the scale do not completely correspond to gradation, as is usually the case. The negative labels include both tweets that do not need fact-checking (the label ``No, no need to check'') and those worth fact-checking but not requiring experts' attention (the label ``No, too trivial to check''). This distinction probably creates ambiguity for the model, as demonstrated by poor predictive accuracy and calibration.
% On the NEWS dataset, the CW task essentially amounts to topic classification. It is unclear how the model should handle sentences unrelated to the provided topics. The authors in the corresponding research paper report performance scores with F1 not exceeding .7, even for annotators.

\section*{Risks}
Although we intend to combat the spread of disinformation with this work, there is still a potential for misuse. The prompts and insights reported in this work could potentially be used to create disinformative claims adapted to make their detection more difficult. A big challenge of disinformation detection is the growing use of generative models for creating disinformative claims. The prompts provided in this work could be reverted for generative purposes, achieving the exact opposite effect than what our work aims to achieve.

\section*{Acknowledgments}
This research was supported by the Adria Digital Media Observatory (ADMO) project, which is part of EDMO, EU’s largest interdisciplinary network for countering disinformation.

% Entries for the entire Anthology, followed by custom entries
\bibliography{anthology,custom}

\appendix

\section{Dataset Information}
\label{sec:dataset_info}
In this section, we provide details on the datasets used in our experiments.

\subsection{Test set selection}
Here, we provide details on the test set selection for each dataset. Furthermore, we state which set the authors used for evaluation and whether the results can be comparable.

\paragraph{ClaimBuster.} The dataset does not have an explicit test set. The authors instead used 4-fold cross-validation on different-sized subsets during their experiments (4,000, 8,000 ... 20,000). However, a high-quality \emph{groundtruth} set is available in the dataset. It contains 1,032 samples that experts agreed on and was used for screening during annotation. Also, all the test sets the authors used contain the screening sentences. For the quality of labels and to have somewhat comparable results to the authors, we selected the \emph{groundtruth} set for experiments.

\paragraph{CLEF.} The dataset consists of both a \emph{dev} and a \emph{test} set. Since the \emph{test} set was used to evaluate teams participating in the CLEF CheckThat! the challenge, we opted to do our experiments on this set to compare to the metrics of the best-submitted solution.

\paragraph{EnvironmentalClaims.} The dataset contains both a \emph{dev} and \emph{test} set of equal size, whereas the original work publishes metrics on both sets separately. We selected the \emph{test} set for our experiments.

\paragraph{NewsClaims.} The dataset provides both a \emph{dev} and a \emph{test} set; however, the disclosed sets contain only positive instances. The complete dataset consists of around 10\% of positive instances, with a high number of low-quality negative instances created by errors in sentencizing and filtering -- instances containing only names, dates, links. The dataset also contains duplicate instances, also in the set of positives. To create a viable subset and avoid high costs during inference, we sampled the negative instances from a normal distribution with the parameters fitted to the length of the instances. We chose to sample the same number of instances as there are positives without duplicates, creating a higher baseline.

\paragraph{PoliClaim.} The dataset provides an explicit \emph{test} set consisting of both gold labels and labels resulting from inference on 4 political speeches. To be able to compare results, we opted to use the complete \emph{test} set.

% \subsection{Original metrics}
% \label{sec:benchmarks}

% \paragraph{ClaimBuster.} As previously mentioned, the authors used 4-fold cross-validation on different-sized subsets during their experiments (4,000, 8,000 ... 20,000). They evaluate using $f_wavg$ with the highest score of .818. Our highest scores on $f_wavg$ are .933 on \emph{gpt-4-turbo} and .906 on \emph{gpt-3.5-turbo}, which is a significant improvement. The authors also evaluate ranking, where our results improve on P@k.

% \paragraph{CLEF.} The best results of the Task 1 on CLEF CheckThat!2022 were accuracy of .761 for claim detection, and the F1 of .698 on check-worthiness detection. While our approach underperforms for check-worthiness detection (F1 of .583), it achieves higher accuracy for claim detection (.776 on Level V2).

% \paragraph{EnvironmentalClaims.} The authors report F1 on the test set, the highest achieved is .849. Our approach achieves .773 for Level V1.

% \paragraph{NewsClaims.} The authors report F1 on the whole dataset. However, it is inconclusive whether it is evaluated on the binary or multiclass labels. They report the highest F1 of .309, which our approach surpasses on the subset we selected, with F1 of .583.

% \paragraph{PoliClaim.} The authors evaluate on accuracy. On the \emph{test} set, they achieve an accuracy of .764 on \emph{gpt-3.5} and .862 on \emph{gpt-4}. Our approach achieves the same maximum accuracy using \emph{gpt-3.5-turbo} with prompt Level V3, but lower accuracy for \emph{gpt-4}.

\subsection{Context information}
\label{sec:context_info}

\paragraph{ClaimBuster.} During the annotation of the ClaimBuster dataset, 4 preceding statements could be viewed with an extra button, which was used in 14\% of all cases. Since the dataset covers presidential debates with multiple speakers, including the moderator and audience questioners, it is not completely clear how the speakers were differentiated in the provided preceding sentences. Therefore, we selected the method of differentiating the speakers arbitrarily -- 'A' was used for the speaker of the statement that is meant to be annotated, and 'B' for the opposing speaker.

\paragraph{EnvironmentalClaims.} No additional contextual or co-textual information was provided in the dataset. The annotators were not shown any co-text during annotations. The authors considered annotating whole paragraphs instead of sentence-level annotation but decided against it due to time and budget constraints. 

\paragraph{PoliClaim.} The annotators were provided with the preceding and following sentences of the one they are annotating. Since there is only one speaker (as opposed to ClaimBuster, which covers debates), there is no need to differentiate the speaker to minimize confusion in prompts. In annotation guidelines, context was explicitly mentioned and clarified in examples. In our experiments, we used two versions of the prompts -- one mentioning context for experiments with co-text expansion and one without the mention of context used when only one sentence from the speech is provided. The two alternatives are shown in \ref{sec:prompts}.

\paragraph{CLEF.} The dataset consists of tweets covering COVID-19 topics. For the check-worthiness task, annotators were shown metadata such as time, account, number of likes and reposts. However, this information is not readily available in the dataset and requires crawling the tweets to obtain it. It was also not available in the dataset of the CLEF2022 CheckThat! Challenge, which was derived from the original dataset. Since we wanted to make our effort comparable to alternative methods used in the competition, we did not opt for crawling the tweets to acquire metadata.  

\paragraph{NewsClaims.} The research paper introducing the dataset has inconsistencies regarding the co-text provided to annotators. While it is stated in the paper that whole articles are provided for co-text, in the screenshot of the annotation platform, only three preceding and following sentences were provided. Regarding context, the work emphasizes the importance of metadata such as claim object, speaker and span, and provides that data for positive instances (sentences containing claims related to 4 specified COVID-19 subtopics). The effort of annotating the claims with metadata is worthwhile, however we decided against using it in inference since no such data is available for negative instances.

\section{Model Information}
\label{sec:models}
For OpenAI models, we use \emph{gpt-3.5-turbo-0125} and \emph{gpt-4-0125-preview}. We use a temperature of 0 for all experiments. To get confidence, we use \emph{logprobs} and \emph{n\_probs=5}, to account for the target labels ending up as less probable tokens. We use a random seed of 42 in all experiments, to avoid stochastic answers as much as possible. The run was executed once per model and prompt variant. Inference was done through the OpenAI API. GPU hours are hard to estimate.

We use Llama3 8B Instruct for experiments on open-source models. It is the only smaller open-source model from the ones we tested compliant with provided labels. The experiments took 10 GPU hours on 2x GeForce RTX 2080 Ti. We use greedy decoding and run once per model and prompt variant.
Initial experiments were done on \emph{neural-chat:7b-v3.3-q5\_K\_M} and \emph{mistral:7b-instruct-v0.2-q5\_K\_M}. A total of 5 GPU hours was used.

For BERT, we use the base model \emph{bert-base-uncased}. We train the model for five epochs with a batch size of 16, a learning rate of 2e-5 and weight decay of 0.01. We keep the best model across epochs.

\section{Calibration}
In this section, the $ECE$ per prompt verbosity level is shown for all models in Table~\ref{tab:ece_all}. The $ECE$ is calculated with the parameters $n_bins=10$ and $norm=l1$.

\begin{table*}
\centering
    {
    \small
        \begin{tabular}{@{}lcccccccc@{}}
            \toprule
             & & \multicolumn{2}{c}{\textbf{CB}} & \multicolumn{2}{c}{\textbf{CLEF}} & \textbf{ENV} & \textbf{NEWS} & \textbf{POLI} \\ 
            \cmidrule(lr){3-4}\cmidrule(lr){5-6}\cmidrule(lr){7-7}\cmidrule(lr){8-8} \cmidrule(lr){9-9}
            & & CD & CW & CD & CW & CD & CW & CW \\
            \midrule
            \multirow{4}{*}{gpt-4-turbo} & V0 & .094 & .068 & .259 & .601 & .231 & .322 &.142 \\ 
            & V1 & .050 & .047 & .196 & .391 & .119 & .210 &.271 \\ 
            & V2 & .043 & .039 & .194 & .352 & .127 & .277 &.373 \\ 
            & V3 & .039 & .032 & .222 & .367 & .150 & .194 &.348 \\
            \midrule
            \multirow{4}{*}{gpt-3.5-turbo} & V0 & .033 & .068 & .212 & .359 & .189 & .246 & .257\\ 
            & V1 & .323 & .085 & .386 & .609 & .088 & .260 & .229 \\ 
            & V2 & .103 & .071 & .279 & .560 & .097 & .280 & .327 \\ 
            & V3 & .061 & .050 & .285 & .646 & .100 & .379 & .196 \\
            \midrule
            \multirow{4}{*}{Llama3 8B} & V0 & .218 & .126 & .307 & .611 & .286 & .314 & .223 \\ 
            & V1 & .607 & .218 & .244 & .723 & .114 & .228 & .172 \\ 
            & V2 & .184 & .135 & .241 & .687 & .102 & .229 & .321 \\ 
            & V3 & .231 & .259 & .241 & .686 & .134 & .214 & .379 \\
            \bottomrule
        \end{tabular}}
\caption{ECE score by prompt level per dataset for \emph{gpt-4-turbo}. 'CD' and 'CW' mark claim detection and claim check-worthiness detection, respectively, while 'V0' marks the score for the naive baseline}
\label{tab:ece_all}
\end{table*}

\section{Complete prompts}
\label{sec:prompts}
This section provides the complete prompts used in our experiments. The instructions were given in system prompts, while the instances were in user prompts. The added context information is also appended to user prompts.

For each dataset, the three prompt levels are shown, with the content expanded in relation to the previous level highlighted. To visually separate the levels, Level V2 is highlighted in yellow, while Level V3 is highlighted in pink.

For CLEF, two alternative prompts are given, since for CD and CW different annotation guidelines were used. For POLI, parts of the Level V2 and Level V3 prompts regarding surrounding sentences are either provided or not, based on whether context expansion is used (surrounding sentences are given in prompts C1 and C3). Those parts are highlighted in blue.

\paragraph{User prompts.} The user prompts were based on how the instance was reffered to in the corresponding annotation gudelines. The instances are surrounded with HTML tags. The same is done for context expansion on CB and POLI.

\begin{table*}
    \centering
    \begin{adjustbox}{width=\textwidth}
    {\small
    \begin{tabular}{c c} 
     \toprule
     \textbf{Level} & \textbf{Prompt} \\ 
     \midrule
    V1 & \parbox{\textwidth}{\myfont{Categorize the <sentence> spoken in the presidential debates into one of three categories: Non-Factual Sentence (NFS), Unimportant Factual Sentence (UFS) or Check-worthy Factual Sentence (CFS). Use only one of the three labels (NFS, UFS or CFS), do not provide any additional explanation.}} \\
    \midrule
     V2 & \parbox{\textwidth}{
     \myfont{Categorize the <sentence> spoken in the presidential debates into three categories:
Non-Factual Sentence (NFS): \hl{Subjective sentences (opinions, beliefs, declarations) and many questions fall under this category. These sentences do not contain any factual claim.}

Unimportant Factual Sentence (UFS): \hl{These are factual claims but not check-worthy. The general public will not be interested in knowing whether these sentences are true or false. Fact-checkers do not find these sentences as important for checking.}

Check-worthy Factual Sentence (CFS): \hl{They contain factual claims and the general public will be interested in knowing whether the claims are true. Journalists look for these type of claims for fact-checking.}

Use only one of the three labels (NFS, UFS and CFS), do not provide any additional explanation.}} \\
    \midrule
 V3 & \parbox{\textwidth}{
  \sethlcolor{pink}
  \myfont{Categorize the <sentence> spoken in the presidential debates into three categories:

Non-Factual Sentence (NFS): Subjective sentences (opinions, beliefs, declarations) and many questions fall under this category. These sentences do not contain any factual claim. \hl{Here are two such examples. ”But I think it’s time to talk about the future.“ “You remember the last time you said that?”}
Unimportant Factual Sentence (UFS): These are factual claims but not check-worthy. The general public will not be interested in knowing whether these sentences are true or false. Fact-checkers do not find these sentences as important for checking. \hl{Some examples are as follows. “Next Tuesday is Election day.” “Two days ago we ate lunch at a restaurant.”}
Check-worthy Factual Sentence (CFS): They contain factual claims and the general public will be interested in knowing whether the claims are true. Journalists look for these type of claims for fact-checking. \hl{Some examples are: “He voted against the first Gulf War.” “Over a million and a quarter Americans are HIV-positive.”}

Use only one of the three labels (NFS, UFS and CFS), do not provide any additional explanation. }}\\
     \bottomrule
    \end{tabular}}
    \end{adjustbox}
    \captionof{table}{System prompts used for inference on the ClaimBuster dataset.}
\label{tab:CB}
\end{table*}

\begin{table*}
    \centering
    \begin{adjustbox}{width=\textwidth}
    {\small
    \begin{tabular}{c c} 
     \toprule
     \textbf{Level} & \textbf{Prompt} \\ 
     \midrule
    V1 & \parbox{\textwidth}{\myfont{Your task is to label the <sentence>. The information I need is whether it is an environmental claim. A broad definition for such a claim is given by the European Commission: Environmental claims refer to the practice of suggesting or otherwise creating the impression that a product or a service is environmentally friendly (i.e., it has a positive impact on the environment) or is less damaging to the environment than competing goods or services. Answer only with Yes or No.}} \\
    \midrule
    V2 & \parbox{\textwidth}{\myfont{Your task is to label the <sentence>. The information I need is whether it is an environmental claim. A broad definition for such a claim is given by the European Commission: Environmental claims refer to the practice of suggesting or otherwise creating the impression that a product or a service is environmentally friendly (i.e., it has a positive impact on the environment) or is less damaging to the environment than competing goods or services. 
\hl{General principles:
You will be presented with a <sentence> and have to decide whether the <sentence> contains an explicit environmental claim. Do not rely on implicit assumptions when you decide on the label. Base your decision on the information that is available within the sentence. However, if a sentence contains an abbreviation, you could consider the meaning of the abbreviation before assigning the label. In case a sentence is too technical/complicated and thus not easily understandable, it usually does not suggest to the average consumer that a product or a service is environmentally friendly and thus can be rejected. Likewise, if a sentence is not specific about having an environmental impact for a product or service, it can be rejected.}
Answer only with Yes or No.}} \\
    \midrule
    V3 & \parbox{\textwidth}{\sethlcolor{pink}
    \myfont{Your task is to label the <sentence>. The information I need is whether it is an environmental claim. A broad definition for such a claim is given by the European Commission: Environmental claims refer to the practice of suggesting or otherwise creating the impression that a product or a service is environmentally friendly (i.e., it has a positive impact on the environment) or is less damaging to the environment than competing goods or services. 
General principles:
You will be presented with a sentence and have to decide whether the sentence contains an explicit environmental claim. Do not rely on implicit assumptions when you decide on the label. Base your decision on the information that is available within the sentence. However, if a sentence contains an abbreviation, you could consider the meaning of the abbreviation before assigning the label. In case a sentence is too technical/complicated and thus not easily understandable, it usually does not suggest to the average consumer that a product or a service is environmentally friendly and thus can be rejected. Likewise, if a sentence is not specific about having an environmental impact for a product or service, it can be rejected.

\hl{Examples:
<sentence>: Farmers who operate under this scheme are required to dedicate 10\% of their land to wildlife preservation.
Label: Yes
Explanation: Environmental scheme with details on implementation.

<sentence>: UPM Biofuels is developing a new feedstock concept by growing Brassica Carinata as a sequential crop in South America.
Label: No
Explanation: Sentence context would be required to understand whether it is a claim.}

Answer only with Yes or No, don't provide any additional explanation.}} \\

     \bottomrule
    \end{tabular}}
    \end{adjustbox}
    \captionof{table}{System prompts used for inference on the EnvironmentalClaims dataset.}
\label{tab:ENV}
\end{table*}

\begin{table*}
    \centering
    \begin{adjustbox}{width=\textwidth}
    {\small
    \begin{tabular}{c c} 
     \toprule
     \textbf{Level} & \textbf{Prompt} \\ 
     \midrule
    V1 & \parbox{\textwidth}{\myfont{A verifiable factual claim is a sentence claiming that something is true, and this can be verified using factual, verifiable information such as statistics, specific examples, or personal testimony.
    Does the <tweet> contain a verifiable factual claim? Answer only with Yes or No, don't provide any additional explanation.}} \\
    \midrule
     V2 & \parbox{\textwidth}{
     \myfont{A verifiable factual claim is a sentence claiming that something is true, and this can be verified using factual, verifiable information such as statistics, specific examples, or personal testimony.

\hl{Factual claims include the following:
Stating a definition;
Mentioning quantity in the present or the past;
Making a verifiable prediction about the future;
Reference to laws, procedures, and rules of operation;
References to images or videos (e.g., "This is a video showing a hospital in Spain.'');
Statements about correlations or causations. Such correlation and causation needs to be explicit, i.e., sentences like "This is why the beaches haven't closed in Florida.'' is not a claim because it does not say why explicitly, thus it is not verifiable.

Tweets containing personal opinions and preferences are not factual claims.

Note: if a tweet is composed of multiple sentences or clauses, at least one full sentence or clause needs to be a claim in order for the tweet to contain a factual claim. If a claim exist in a sub-sentence or sub-clause then tweet is not considered to have a factual claim. For example, "My new favorite thing is Italian mayors and regional presidents LOSING IT at people violating quarantine'' is not a claim, however, it is an opinion. Moreover, if we consider "Italian mayors and regional presidents LOSING IT at people violating quarantine'' it would be a claim. In addition, when answering this question, annotator should not open the tweet URL.}

Does the <tweet> contain a verifiable factual claim?
Answer only with Yes or No.}} \\
    \midrule
 V3 & \parbox{\textwidth}{
  \sethlcolor{pink}
  \myfont{A verifiable factual claim is a sentence claiming that something is true, and this can be verified using factual, verifiable information such as statistics, specific examples, or personal testimony.

Factual claims include the following:
Stating a definition;
Mentioning quantity in the present or the past;
Making a verifiable prediction about the future;
Reference to laws, procedures, and rules of operation;
References to images or videos (e.g., "This is a video showing a hospital in Spain.'');
Statements about correlations or causations. Such correlation and causation needs to be explicit, i.e., sentences like "This is why the beaches haven't closed in Florida.'' is not a claim because it does not say why explicitly, thus it is not verifiable.

Tweets containing personal opinions and preferences are not factual claims.

Note: if a tweet is composed of multiple sentences or clauses, at least one full sentence or clause needs to be a claim in order for the tweet to contain a factual claim. If a claim exist in a sub-sentence or sub-clause then tweet is not considered to have a factual claim. For example, "My new favorite thing is Italian mayors and regional presidents LOSING IT at people violating quarantine'' is not a claim, however, it is an opinion. Moreover, if we consider "Italian mayors and regional presidents LOSING IT at people violating quarantine'' it would be a claim. In addition, when answering this question, annotator should not open the tweet URL.

Does the <tweet> contain a verifiable factual claim?
Answer only with Yes or No. 

\hl{Examples:
Tweet: Please don't take hydroxychloroquine (Plaquenil) plus Azithromycin for \#COVID19 UNLESS your doctor prescribes it. Both drugs affect the QT interval of your heart and can lead to arrhythmias and sudden death, especially if you are taking other meds or have a heart condition.
Label: Yes
Explanation: There is a claim in the text.
Tweet: Saw this on Facebook today and it’s a must read for all those idiots clearing the shelves \#coronavirus \#toiletpapercrisis \#auspol
Label: No
Explanation: There is no claim in the text.}

Answer only with Yes or No, don't provide any additional explanation.}}\\
     \bottomrule
    \end{tabular}}
    \end{adjustbox}
    \captionof{table}{System prompts used for inference on the CLEF dataset for claim detection.}
\label{tab:CLEF_CD}
\end{table*}

\begin{table*}
    \centering
    \begin{adjustbox}{width=\textwidth}
    {\small
    \begin{tabular}{c c} 
     \toprule
     \textbf{Level} & \textbf{Prompt} \\ 
     \midrule
    V1 & \parbox{\textwidth}{\myfont{It is important that a verifiable factual check-worthy claim be verified by a professional fact-checker, as the claim may cause harm to society, specific person(s), company(s), product(s), or some government entities. However, not all factual claims are important or worth fact-checking by a professional fact-checker, as this very time-consuming. Do you think that a professional fact-checker should verify the claim in the <tweet>?
Labels:
No, no need to check; No, too trivial to check; Yes, not urgent; Yes, very urgent. 

Decide on one label. Then, answer only with Yes or No.}} \\
    \midrule
     V2 & \parbox{\textwidth}{
     \myfont{It is important that a verifiable factual check-worthy claim be verified by a professional fact-checker, as the claim may cause harm to society, specific person(s), company(s), product(s), or some government entities. However, not all factual claims are important or worth fact-checking by a professional fact-checker, as this very time-consuming. Do you think that a professional fact-checker should verify the claim in the <tweet>?
Labels:
No, no need to check: \hl{the tweet does not need to be fact-checked, e.g., be- cause it is not interesting, a joke, or does not contain any claim.}
No, too trivial to check: \hl{the tweet is worth fact-checking, however, this does not require a professional fact-checker, i.e., a non-expert might be able to fact-check the claim. For example, one can verify the information using reliable sources such as the official website of the WHO, etc. An example of a claim is as follows: “The GDP of the USA grew by 50\% last year.”}
Yes, not urgent: \hl{the tweet should be fact-checked by a professional fact-checker, however, this is not urgent or critical;}
Yes, very urgent: \hl{the tweet can cause immediate harm to a large number of people; therefore, it should be verified as soon as possible by a professional fact-checker;}

Decide on one label. Then, answer only with Yes or No.}} \\
    \midrule
 V3 & \parbox{\textwidth}{
  \sethlcolor{pink}
  \myfont{It is important to verify a factual claim by a professional fact-checker, which can cause harm to the society, specific person(s), company(s), product(s) or government entities. However, not all factual claims are important or worthwhile to be fact-checked by a professional fact-checker as it is a time-consuming procedure. Do you think that a professional fact-checker should verify the claim in the <tweet>?
Labels:
No, no need to check: the tweet does not need to be fact-checked, e.g., be- cause it is not interesting, a joke, or does not contain any claim.
No, too trivial to check: the tweet is worth fact-checking, however, this does not require a professional fact-checker, i.e., a non-expert might be able to fact-check the claim. For example, one can verify the information using reliable sources such as the official website of the WHO, etc. An example of a claim is as follows: “The GDP of the USA grew by 50% last year.”
Yes, not urgent: the tweet should be fact-checked by a professional fact-checker, however, this is not urgent or critical;
Yes, very urgent: the tweet can cause immediate harm to a large number of people; therefore, it should be verified as soon as possible by a professional fact-checker;

\hl{Examples:
Tweet: Wash your hands like you've been chopping jalapeños and need to change a contact lens'' says BC Public Health Officer Dr. Bonnie Henry re. ways to protect against \#coronavirus \#Covid\_19
Label: Yes, not urgent
Explanation: Overall it is less important for a professional fact-checker to verify this information. The statement does not harm anyone. The truth value of whether the official said the statement is not important. Also it appears that washing hands is very important to protect oneself from the virus.
Tweet: ALERT! The corona virus can be spread through internationaly printed albums. If you have any albums at home, put on some gloves, put all the albums in a box and put it outside the front door tonight. I'm collecting all the boxes tonight for safety. Think of your health.
Label: No, no need to check
Explanation: This is joke and no need to check by a professional fact checker.}

Decide on one label. Then, answer only with Yes or No.}}\\
     \bottomrule
    \end{tabular}}
    \end{adjustbox}
    \captionof{table}{System prompts used for inference on the CLEF dataset for claim check-worthiness detection.}
\label{tab:CLEF_CW}
\end{table*}

\begin{table*}
    \centering
    \begin{adjustbox}{width=\textwidth}
    {\small
    \begin{tabular}{c c} 
     \toprule
     \textbf{Level} & \textbf{Prompt} \\ 
     \midrule
    V1 & \parbox{\textwidth}{\myfont{The task is to select verifiable statements from political speeches for fact-checking. Given a \textless statement \textgreater from a political speech, answer the question. 
    Does the \textless statement \textgreater explicitly present any verifiable factual information?
    Answer with A, B or C only. A - Yes, B - Maybe, C - No.}} \\
    \midrule
     V2 & \parbox{\textwidth}{
     \myfont{The task is to select verifiable statements from political speeches for fact-checking. Given a \textless statement \textgreater from a political speech, answer the question following the guidelines.  
\hl{Definitions and guidelines:
Fact: A fact is a statement or assertion that can be objectively verified as true or false based on empirical evidence or reality.
Opinion: An opinion is a judgment based on facts, an attempt to draw a reasonable conclusion from factual evidence. While the underlying facts can be verified, the derived opinion remains subjective and is not universally verifiable.} \sethlcolor{cyan}\hl{
Context: Make sure to consider a small context of the target statement (the previous and next sentence) when annotating. Some statements require context to understand the meaning.} \sethlcolor{yellow}\hl{
Factual claim: A factual claim is a statement that explicitly presents some verifiable facts. Statements with subjective components like opinions can also be factual claims if they explicitly present objectively verifiable facts.
Opinion with Facts: Opinions can also be based on factual information.
When does an opinion explicitly present a fact: Many opinions are more or less based on some factual information. However, some facts are explicitly presented by the speakers, while others are not.
What is verifiable: The verifiability of the factual information depends on how specific it is. If there is enough specific information to guide a general fact-checker in checking it, the factual information is verifiable. Otherwise, it is not verifiable.}

The question: Does the \textless statement \textgreater explicitly present any verifiable factual information?
Answer with A, B or C only.
\hl{A - Yes, the statement contains factual information with enough specific details that a fact-checker knows how to verify it. E.g., Birmingham is small in population compared to London.
B - Maybe, the statement seems to contain some factual information. However, there are certain ambiguities (e.g., lack of specificity) making it hard to determine the verifiability. E.g., Birmingham is small compared to London. (lack of details about what standard Birmingham is small)
C - No, the statement contains no verifiable factual information. Even if there is some, it is clearly unverifiable. E.g., Birmingham is small.}}}\\
     \bottomrule
    \end{tabular}}
    \end{adjustbox}
    \captionof{table}{System prompts of Level V1 and Level V2 used for inference on the PoliClaim dataset for claim check-worthiness detection. The blue highlight shows instructions for regarding context.}
\label{tab:POLI1}
\end{table*}

\begin{table*}
    \centering
    \begin{adjustbox}{width=\textwidth}
    {\small
    \begin{tabular}{c c} 
     \toprule
     \textbf{Level} & \textbf{Prompt} \\ 
     \midrule
 V3 & \parbox{\textwidth}{
  \sethlcolor{pink}
  \myfont{The task is to select verifiable statements from political speeches for fact-checking. Given a statement from a political speech and its context, answer the question following the guidelines. 
\hl{Definitions and guidelines:
Fact: A fact is a statement or assertion that can be objectively verified as true or false based on empirical evidence or reality.
Opinion: An opinion is a judgment based on facts, an attempt to draw a reasonable conclusion from factual evidence. While the underlying facts can be verified, the derived opinion remains subjective and is not universally verifiable.
Factual claim: A factual claim is a statement that explicitly presents some verifiable facts. Statements with subjective components like opinions can also be factual claims if they explicitly present objectively verifiable facts.}
\sethlcolor{cyan}\hl{
Context: Make sure to consider a small context of the target statement (the previous and next sentence) when annotating. Some statements require context to understand the meaning. For example:
E1. “... Just consider what we did last year for the middle class in California, sending 12 billion dollars back – the largest state tax rebate in American history. \textless statement \textgreater But we didn’t stop there. \textless \_statement \textgreater We raised the minimum wage. We increased paid sick leave. Provided more paid family leave. Expanded child care to help working parents...” Without the context, the sentence marked with \textless statement \textgreater seems an incomplete sentence. With the context, we know the speaker is claiming a bunch of verifiable achievements of their administration.
E2. “... When I first stood before this chamber three years ago, I declared war on criminals and asked for the Legislature to repeal and replace the catch-and-release policies in SB 91. \textless statement \textgreater With the help of many of you, we got it done. <\_statement> Policies do matter. We’ve seen our overall crime rate decline by 10 percent in 2019 and another 18.5 percent in 2020! ...” The part marked with <statement> claims that the policies against crimes have been “done”, which is verifiable. It needs context to understand it.}

\sethlcolor{pink}\hl{
Opinion with Facts: Opinions can also be based on factual information. For example:
E1. “I am proud to report that on top of the local improvements, the state has administered projects in almost all 67 counties already, and like I said, we’ve only just begun.” The speaker’s “proud of” is a subjective opinion. However, the content of pride (administered projects) is factual information.
E2. “I first want to thank my wife of 34 years, First Lady Rose Dunleavy.” The speaker expresses their thankfulness to their wife. However, there is factual information about the first lady’s name and the length of their marriage.

When does an opinion explicitly present a fact: Many opinions are more or less based on some factual information. However, some facts are explicitly presented by the speakers, while others are not. Explicit presentation means the fact is directly entailed by the opinion without extrapolation:
E1. “The pizza is delicious.” This opinion seems to be based on the fact that “pizza is a kind of food”. However, this fact is not explicitly presented.
E2. “I first want to thank my wife of 34 years, First Lady Rose Dunleavy.” The name of the speaker’s wife and their year of marriage are explicitly presented.

What is verifiable: The verifiability of the factual information depends on how specific it is. If there is enough specific information to guide a general fact-checker in checking it, the factual information is verifiable. Otherwise, it is not verifiable. 
E1. “Birmingham is small.” is not verifiable because it lacks any specific information for determining veracity. It leans more toward subjective opinion.
E2. “Birmingham is small, compared to London” is more verifiable than E1. A fact-checker can retrieve the city size, population size ...etc., of London and Birmingham to compare them. However, what to compare to prove Birmingham’s “small” is not specific enough.
E3. “Birmingham is small in population size, compared to London” is more verifiable than E1 and E2. A fact-checker now knows it is exactly the population size to be compared.}

The question: 
Does the <statement> explicitly present any verifiable factual information?
Answer with A, B or C only.
A - Yes, the statement contains factual information with enough specific details that a fact-checker knows how to verify it. E.g., Birmingham is small in population compared to London.
B - Maybe, Maybe, the statement seems to contain some factual information. However, there are certain ambiguities (e.g., lack of specificity) making it hard to determine the verifiability. E.g., Birmingham is small compared to London. (lack of details about what standard Birmingham is small)
C - No, the statement contains no verifiable factual information. Even if there is some, it is clearly unverifiable. E.g., Birmingham is small.}} \\
     \bottomrule
    \end{tabular}}
    \end{adjustbox}
    \captionof{table}{System prompts of Level V3 used for inference on the PoliClaim dataset for claim check-worthiness detection. The blue highlight shows instructions for regarding context.}
\label{tab:POLI2}
\end{table*}

\end{document}